# General Belief Measures

Emil Weydert
Max-Planck-Institute for Computer Science
Im Stadtwald, D-66123 Saarbrücken, Germany
emil@mpi-sb.mpg.de

## Abstract

Probability measures by themselves are known to be inappropriate for modeling the dynamics of plain belief and their excessively strong measurability constraints make them unsuitable for some representational tasks, e.g. in the context of first-order knowledge. In this paper, we are therefore going to look for possible alternatives and extensions. We begin by delimiting the general area of interest, proposing a minimal list of assumptions to be satisfied by any reasonable quasi-probabilistic valuation concept. Within this framework, we investigate two particularly interesting kinds of quasi-measures which are not or much less affected by the traditional problems.
• Ranking measures, which generalize Spohn-type and possibility measures.
• Cumulative measures, which combine the probabilistic and the ranking philosophy, allowing thereby a fine-grained account of static and dynamic belief.

## 1 INTRODUCTION

The successful acting of cognitive agents in complex, opaque and dynamic worlds depends on their ability to manipulate huge amounts of incomplete and uncertain information in a reasonable way. The representation of soft and partial knowledge together with the formalization of the corresponding reasoning patterns are therefore major issues in artificial intelligence. Very roughly, we can distinguish the following approaches to the modeling of belief, i.e. of uncertain or revisable knowledge :

• *Strictly qualitative.*
Belief states are represented by specific prioritized sets of sentences (*syntactic*) [Weydert 92, Nebel 92], or by some preference relation on possible worlds structures inducing an epistemic-entrenchment-like ordering on propositions (*semantic*) [Gärdenfors, Makinson 88, 91].

• *Strictly quantitative.*
Belief states are characterized by classical probabilistic measures [Pearl 88] or by alternative numerical accounts, e.g. belief functions [Shafer 76].

• *Semi-quantitative* or *semi-qualitative*.
Belief states are described by rough qualitative measures assigning orders of magnitude [e.g. Spohn 88].

For most proposals, there are more or less suitable application contexts, but altogether they also exhibit some more or less serious shortcomings. Prioritized sentential belief models are interesting because of their simplicity and the absence of omniscience assumptions, but the lacking semantic foundations and the resulting ad hoc character relativize their usefulness. More generally, strictly qualitative syntax- or semantic-based comparative frameworks are not fine-grained enough to handle the complexities of real-world knowledge. For instance, we cannot easily model belief strength differences or independency constraints and our decision-theoretic considerations have to stay rather rudimentary.

Among the numerical approaches, probability theory is certainly the best investigated and most successful formalism for modeling uncertain propositional knowledge. Even if its naive use might sometimes cause problems, it still appears to be the ideal reference formalism for more sophisticated or special-purpose representations, not to mention the practical importance of the powerful tools it provides. But there are several general problems which have to be addressed, in particular if we want to handle changing or first-order belief states.

• Agents cannot always associate exact numbers - e.g. probabilities reflecting betting intentions and/or statistical knowledge - with propositions. Very often, numerical values are undefined, unknown, irrelevant, cumbersome or untractable, and more transparent, simpler models of reality are required, lowering the computational costs.

• Standard conditionalization procedures in the traditional probabilistic context don't allow us to revise plain belief if we choose to implement it, quite naturally, by assigning (subjective) probability 1.

• The representation of complex forms of (e.g. first-order) knowledge dealing with infinite structures is frequently blocked by unsatisfiable measurability constraints.

- There are intuitive flaws, for instance the gap between impossibility and probability 0 or the nonexistence of uniform distributions on countably infinite sets.

To a certain extent, all these problems can be solved in semi-qualitative quasi-probabilistic formal accounts based on abstract ranking measures. Their characteristic feature is that the value associated with the union of two sets is assumed to be, w.r.t. a given ranking order, just the maximum of the values attributed to the individual sets. This means less commitments by lower precision. Homogeneous countably infinite sets and proper impossibility are no longer barred and every ranking measure can be lifted to the full power set. Related concepts have been used fairly successfully for multiple revisions of plain belief [Spohn 88], default reasoning [Weydert 91, Dubois, Prade 91, Goldszmidt, Pearl 92] and belief modeling [Weydert 94]. Their main disadvantage is that they can only provide a very rough picture, suitable for interpreting defaults but not for a (really) fine-grained decision-theoretic analysis. So, in some sense, quantitative and qualitative formalisms seem to offer complementary features which we might try to combine in a new type of mixed framework, possibly powerful enough for more realistic dynamic models of graded first-order belief.

In this paper, we are going to develop such an integrated approach by considering natural generalizations of probability measures. We shall proceed in four steps. To begin with, we shall delimit the area of interest and single out the minimal conditions for reasonable quasi-probabilistic valuation formalisms. Next, we shall introduce and discuss two particularly interesting subclasses of non-classical quasi-measures. On one hand, we have the ranking measures, a coarse-grained semi-qualitative notion. On the other hand, we have the cumulative measures, a fine-grained extended quantitative-qualitative measure concept, which tries to accumulate the best of both worlds. To conclude, we shall sketch how cumulative measures may be used to model static and dynamic belief.

## 2 QUASI-MEASURES

Traditional probabilistic measure spaces can be seen as triples of the form $(\mathcal{B}, \mathcal{P}, \mathcal{V})$, where $\mathcal{B} = (B, \cup, \cap, -, 0, 1)$ is a $\sigma$-algebra of events, i.e. a boolean algebra closed under countable joins (unions) and meets (intersections) with top $1$ and bottom $0$, $\mathcal{V} = (IR^+, +, \times, 0, 1, <)$ is the positive half of the ordered real number field, and the probability measure $\mathcal{P}$ is a function from B to $IR^+$ s.t. $\mathcal{P}(1) = 1$ and $\mathcal{P}(\cup A_i) = \Sigma \mathcal{P}(A_i)$ for every countable set of $\cap$-disjoint $A_i$ ($\sigma$-additivity). For each such space, there is a corresponding conditional probability measure $\mathcal{P}(\mid) : B \times B^\circ \rightarrow IR^+$, which verifies $\mathcal{P}(A \cap B) = \mathcal{P}(A \mid B) \times \mathcal{P}(B)$ for $B \in B^\circ = \{B \in B \mid \mathcal{P}(B) \neq 0\}$. In the context of $\mathcal{P}$, $A_1 \ldots A_n$ are called (conditionally) independent given $B \in B^\circ$ iff $\mathcal{P}(\cap A_{ij} \mid B) = \mathcal{P}(A_{i1} \mid B) \times \ldots \times \mathcal{P}(A_{is} \mid B)$ for each subsequence $A_{i1} \ldots A_{is}$ of $A_1 \ldots A_n$.

Our first task now will be to see how far we can relax all these requirements without giving up practical relevance for belief valuation or basic features of the probability calculus. In particular, we want to keep reasonable notions of conditioning and independency. On the other hand, our framework should be general enough to cover semi-qualitative Spohn-type formalisms like those introduced for modeling defaults and revision. In the following, we are going to propose a minimal list of assumptions to be satisfied by any structure claiming to support a reasonable form of generalized probabilistic reasoning[1]. Before, let's recall some useful algebraic notions.

$(G, *)$ is called a semi-group iff $G \neq \emptyset$ and $*$ is an associative operation on $G$. In $(G, *)$, we call $e$ neutral or the identity iff for all $x \in G$, $e * x = x * e = x$ and $n$ absorptive iff for all $x \in G$, $x * n = n * x = n$. A semi-group $(G, *)$ is a group iff it has an identity $e$ and for all $x \in G$, there is $y \in G$ s.t. $x * y = y * x = e$. $(G, *, <)$ is called an ordered group iff $(G, *)$ is a group, $<$ is a linear ordering on $G$ and for all $x, x', y \in G$, $x < x'$ implies $x * y < x' * y$).

**Definition 2.1** $(\mathcal{B}, \mathcal{R}, \mathcal{V})$ is a *quasi-measure space* with *event algebra* $\mathcal{B} = (B, \cup, \cap, -, 0, 1)$, *valuation algebra* $\mathcal{V} = (V, \#, \circ, n, e, \ll)$ (generalizing $(IR^+, +, \times, 0, 1, <)$) and *quasi-measure* $\mathcal{R} : B \rightarrow V$ iff the postulates **I.0 - I.9** hold

**I.0 Boolean structure :**
$\mathcal{B} = (B, \cup, \cap, -, 0, 1)$ is a boolean algebra.

The usual propositional connectives are fundamental reasoning tools. Infinitary closures are quite exotic.

**I.1 Valuation function :**
$\mathcal{R} : B \rightarrow V$ is a function with $\mathcal{R}(0) = n$, $\mathcal{R}(1) = e$ and $\mathcal{R}(A \cup B) = \mathcal{R}(A) \# \mathcal{R}(B)$ for $\cap$-disjoint A, B.

By measure intuitions. If $A \cap B = 0$, $\mathcal{R}(A \cup B)$ should only depend on $\mathcal{R}(A)$ and $\mathcal{R}(B)$. To grasp infinitesimal probabilies, we have to give up (full) $\sigma$-additivity because of missing upper bounds in nonstandard analysis.

**I.2 Additive structure :**
$(V, \#)$ is a commutative semi-group with identity $n$.

Induced by the boolean properties of $\cup$ and $0$ together with the **I.1** characterization of $\mathcal{R}$. Strictly speaking, this requirement only concerns the possible values of $\mathcal{R}$, but we are going to extrapolate these algebraic properties to the whole structure.

**I.3 Multiplicative structure :** $(V, \circ)$ is a commutative semi-group with identity $e$ and absorptive $n$.

Based on the boolean properties of $\cap, 0, 1$ and the desire to set $\mathcal{R}(A \cap B) = \mathcal{R}(A) \circ \mathcal{R}(B)$ for intuitively inde-pendent events A, B (∘ is meant to play the role of ×), in particular for A, B ∈ {0, 1}.

---

[1] A different and slightly more opaque generalization of probability theory has also been developed by [Darwiche, Ginsberg 92]. Their account is weaker insofar as it allows partially ordered valuation structures.

**I.4 Distributivity:**
$\forall v, v', w \ (w \circ (v \# v') = (w \circ v) \# (w \circ v'))$.

Backed by the distributivity of ∩ over ∪ and our considerations above.

**I.5 Linearity:**
(V, «) is a linear ordering.

For decision-theoretic reasons and formal convenience. If necessary, we can model partiality by sets of valuations.

**I.6 Additive monotony:**
$\forall v, v', w \ (v \underset{\ll}{} v' \to v \# w \underset{\ll}{} v' \# w)$.

Suggested by the monotonic character of measures. To account for ranking measures, we need the weak version $\underset{\ll}{}$.

**I.7 Multiplicative monotony:**
$\forall v, v', w \ (v \ll v' \ \& \ w \neq n \to v \circ w \ll v' \circ w)$.

Conditioning on an independent, non-empty event should not affect «-relationships.

**I.8 Additive accessibility:**
$\forall v, v' \ (v \underset{\ll}{} v' \to \exists w \ v \# w = v')$

It should be possible to attribute an arbitrary lower value to a proper subvent (valuation freedom). Then, the principle follows from **I.1**.

**I.9 Multiplicative accessibility:**
$\forall v, v' \ (v \underset{\ll}{} v' \to \exists w \ v' \circ w = v)$

This condition is necessary if we accept valuation freedom and want a suitable conditional version of $\mathcal{R}$.

To get a better understanding of these structures, we now consider some interesting easy consequences, e.g. the fact that $\underset{\ll}{}$ or « can be defined from ∘.

**Theorem 2.1** Assuming **I.0 - I.9**, we have,

1. $\forall v \ n \underset{\ll}{} v$,
2. $\forall v, v' \ (v \underset{\ll}{} v' \ \& \ v' \neq n \to \exists^{=1} w \ v' \circ w = v)$,
3. $\forall v, v' \ (v \circ v' = n \ \& \ v' \neq n \to v = n)$,
4. $\forall v, v' \ (v \underset{\ll}{} v' \leftrightarrow \exists w \underset{\ll}{} e \ v' \circ w = v)$,
5. $\forall v, v' \ (n \ll v, v' \underset{\ll}{} e \to n \ll v \circ v' \underset{\ll}{} e)$,
6. $\forall A \in B \ n \underset{\ll}{} \mathcal{R}(A) \underset{\ll}{} e$.

**Proof:**

1. By multiplicative accessibility, $v \ll n$ would impy that there is a w s.t. $v = n \circ w = n$, which is impossible. Using linearity, this gives us $n \underset{\ll}{} v$.

2. Because of multiplicative accessibility, we only have to show uniqueness. Suppose, we had $v \underset{\ll}{} v'$, $v' \neq n$ and $w \neq w'$ s.t. $v' \circ w = v' \circ w' = v$. W.l.o.g., we may assume w « w'. But this would contradict multiplicative monotony.

3. Given $n \underset{\ll}{} v'$ and $v' \circ n = n$, **2.1.2** allows us to infer from $v' \circ v = n$ and $v' \neq n$ that $v = n$.

4. If there is a w s.t. $v' \circ w = v$ and $w \underset{\ll}{} e$, either $v = n \underset{\ll}{} v'$, and we are done, or $n \ll v, v', w$ (by **2.1.1, 2.1.3**). Now, suppose that $v' \ll v$ and consequently $w \ll e$. Then multiplicative monotony would give us $v' \circ w \ll v \circ w \ll v \circ e = v$, which contradicts our assumption. So we must have $v \underset{\ll}{} v'$. On the other hand, if $v \underset{\ll}{} v'$, then multiplicative accessibility guarantees the existence of a w s.t. $v' \circ w = v$. If $e \ll w$, multiplicative monotony gives us either $v' = v' \circ e \ll v' \circ w = v$, which is impossible, or $v \underset{\ll}{} v' = n$. But then, $v' \circ w = n = v$ for $w = n \underset{\ll}{} e$.

5. Multiplicative monotony and $n \ll v, v' \underset{\ll}{} e$ imply that $n = n \circ v' \ll v \circ v' \underset{\ll}{} e \circ v' = v' \underset{\ll}{} e$.

6. By **2.1.1** and additive monotony, $n \underset{\ll}{} \mathcal{R}(A) = n \# \mathcal{R}(A) \underset{\ll}{} \mathcal{R}(-A) \# \mathcal{R}(A) = \mathcal{R}(-A \cup A) = \mathcal{R}(1) = e$.

Now we are able to give an explicit definition of conditioning for quasi-measures.

**Definition 2.2** For every quasi-measure space $(\mathcal{B}, \mathcal{R}, \mathcal{V})$, the associated *conditional quasi-measure* $\mathcal{R}(\ |\ ) : B \times B \to V$ is defined as follows. If $\mathcal{R}(B) \neq n$, then $\mathcal{R}(A\ |\ B)$ is the unique $w \in [n, e]_\ll$ s.t. $\mathcal{R}(A \cap B) = \mathcal{R}(B) \circ w$ (by **2.1.2**). If $\mathcal{R}(B) = n$ then $\mathcal{R}(A\ |\ B) = n$.

We can justify this denotation by the following result.

**Theorem 2.2** If $(\mathcal{B}, \mathcal{R}, \mathcal{V})$ is a quasi-measure space, $B \in B, \mathcal{R}(B) \neq n$ and $\mathcal{R}_B(X) = \mathcal{R}(X\ |\ B)$ for all $X \in B$, then $(\mathcal{B}, \mathcal{R}_B, \mathcal{V})$ is also a quasi-measure space.

**Proof:** We only have to verify **I.1**. Obviously $\mathcal{R}(0\ |\ B) = n, \mathcal{R}(1\ |\ B) = e$. What we still must prove is that $\mathcal{R}(A \cup A'\ |\ B) = \mathcal{R}(A\ |\ B) \# R(A'\ |\ B)$ for ∩-disjoint A, A'. Obviously, this holds for $\mathcal{R}(B) = n$. For $\mathcal{R}(B) \neq n$, by **2.1.2**, it is enough to show that $\mathcal{R}((A \cup A') \cap B) = (\mathcal{R}(A\ |\ B) \# R(A'\ |\ B)) \circ \mathcal{R}(B)$. But $(\mathcal{R}(A\ |\ B) \# R(A'\ |\ B)) \circ \mathcal{R}(B) = (\mathcal{R}(A\ |\ B) \circ \mathcal{R}(B)) \# (R(A'\ |\ B) \circ \mathcal{R}(B)) = \mathcal{R}(A \cap B) \# \mathcal{R}(A' \cap B) = \mathcal{R}((A \cap B) \cup (A' \cap B)) = \mathcal{R}((A \cup A') \cap B)$ by distributivity.

Assuming $n \neq e$, the relevant ordered multiplicative substructure of a given valuation algebra $\mathcal{V}$ is characterized by the restriction of (V, ∘, «) to the interval $]n, e]_\ll$. Set $I(\mathcal{V}) = (]n, e]_\ll, \circ, »)$, » being the converse of «.

**Theorem 2.3** $I(\mathcal{V})$ is the positive half of an ordered commutative group.

**Proof:** Based on our definitions and **2.1.5**, it is possible to show that $I(\mathcal{V})$ is an ordered commutative semi-group with identity and minimum e. Now, exploi-ting **2.1.2** (for inverse uniqueness), we can use a well-known technique - integer construction from the natural numbers - to built an ordered commutative group $(G, \circ', \ll')$ whose positive half is isomorphic to $I(\mathcal{V})$.

One way to distinguish valuation algebras is to consider the interaction patterns they induce between # and «. So, for every valuation context, we would like to know how much we have to add to an abstract quantity to get a bigger one. This amounts to investigate for every valuation algebra $\mathcal{V} = (V, \#, \circ, n, e, \ll)$ the corresponding *additive magnitude ordering* $\ll$, defined by $v \ll w$ iff $w \# v = w$. It is easy to see that $\ll \subseteq \ll$ (by additive monotony) is both transitive (associativity of #) and antisymmetric (commutativity of #), but not necessarily reflexive, e.g. if $\mathcal{V} = (\mathbb{R}^+, +, \times, 0, 1, <)$. Notice however that $n \ll n$. Furthermore, it follows from additive monotony and accessibility that $\ll$ is $\ll$-extendible, i.e. if $v \ll x \ll x' \ll v'$, then we get $v \ll v'$ (by additive accessibility, $v' = x' \# w$ and therefore $v' \ll v \# v' = v \# x' \# w' \ll x \# x' \# w' = x' \# w' = v'$, i.e. $v' \# v = v'$).

In general, $\ll$ is much coarser than $\ll$. Setting $S(w) = \{v \mid v \ll w\}$, by extendibility an initial $\ll$-segment, there are basically three possibilities for each $w$ : (1) $S(w) = \{n\}$, (2) $\{n\} \subset S(w) \subset [n, w]_\ll$, (3) $[n, w]_\ll \subseteq S(w)$. In fact, we are going to show that it is enough to consider $w = e$. In the following, let **SP**, **SH** and **SR** be the principles obtained by stating (1), (2) resp. (3) for $w = e$.

**Theorem 2.4** If $n \neq e$, for $i = 1, 2, 3$, (i) holds for $e$ iff (i) holds for all $x \neq n$.

**Proof**: The right-to-left direction is obvious.

(1) Suppose $S(e) = \{n\}$ and $n \neq x = x \# y$ for some $y \neq n$. W.l.o.g. we may assume $y \ll x$, otherwise, additive monotony would give us $x \ll x \# x \ll x \# y = x$ and we could replace $y$ by $x$. By multiplicative accessibility, there is $v \neq n$ s.t. $y = x \circ v$. Because $e \ll e \# v$ and $x \neq n$, we have $x = x \circ e \ll x \circ (e \# v) = x \# (x \circ v) = x \# y$, which is impossible. Hence, (1) must hold for all $x \neq n$.

(2) Suppose $x \neq n$ and $e = e \# v \ll e \# e$ for some $v$ with $n \ll v \ll e$. Then $x = x \circ e = x \circ (e \# v) \ll x \circ (e \# e) = x \# x$ and $n = x \circ n \ll x \circ v \ll x \circ e = x$. In other words, there is $y = x \circ v$ s.t. $n \ll y \ll x$ and $x = x \circ (e \# v) = x \circ e \# x \circ v = x \# y \ll x \# x$. Consequently, (2) is verified for all $x \neq n$.

(3) From $e = e \# e$, we get $x = x \circ (e \# e) = x \# x$. That is, (3) is valid for all $x$.

It seems natural to assume that $\ll$ should define a linear hierarchy of additive magnitudes. This amounts to require that $\ll$ should be a (transitive) modular ordering, i.e. for each $w, v \ll v'$ should imply $w \ll v'$ or $v \ll w$.

**Definition 2.3** A valuation algebra $\mathcal{V} = (V, \#, \circ, n, e, \ll)$ is called *hierarchical* iff $\ll$ is modular.

Standard (probabilistic) and non-standard real valuation algebras - based on the ordered real number field $(\mathbb{R}^+, +, \times, 0, 1, <)$ resp. one of its elementary extensions including infinitesimals - will verify **SP**. Here, # is a very sensitive connective and adding non-zero quantities always makes a difference. In this context, $\ll$ is just $\{n\} \times V \setminus \{n\}$ and therefore modular. That is, these valuation algebras are trivially hierarchical. The remaining two configuration types **SH** and **SR** reflect different qualitative-quantitative philosophies, which we are going to investigate now.

## 3 RANKING MEASURES

The principle **SR**, which is equivalent to idempotence for #, characterizes what might be called pure semi-quantitative quasi-probabilistic accounts. Adopting this condition considerably simplifies our valuation structure. First, it allows us to identify # with $\max_\ll$, because for $n \ll x \ll y$, we have $y \ll x \# y \ll y \# y = y = \max_\ll\{x, y\}$. Together with **2.1.5** and the absorptive character of $n$, this guarantees that $[n, e]_\ll$ is closed for # and $\circ$. Furthermore, **2.1.6** gives us $\mathcal{R}"B \subseteq [n, e]_\ll$. Therefore, w.l.o.g., we may concentrate on **SR**-type valuation algebras of the form $([n, e]_\ll, \max_\ll, \circ, n, e, \ll)$. From **2.3** we know that this structure can be completely described by the positive half of an ordered commutative group which has been topped by an absorptive $n$. The converse of the resulting extended order is just our valuation ordering $\ll$. Another interesting consequence is that we can now drop the disjointness condition in **I.1**. Because we want to ensure that $\mathcal{R}(A) = n$ encodes real impossibility, unions of impossible events should not be allowed to embrace a possible one. All this suggests the following definition extending previous proposals in [Weydert 91, 92].

**Definition 3.1** $(\mathcal{B}, \mathcal{R}, \mathcal{V})$ is called a *ranking-measure space* iff

1. $\mathcal{B} = (B, \cup, \cap, -, 0, 1)$ is a boolean algebra.
2. $\mathcal{V} = (V, \circ, \ll)$ (*ranking algebra*) is s.t.
   - $(V \setminus \{n\}, \circ, \gg)$ is the positive half of an ordered commutative group with identity $e$,
   - $n$ is $\ll$-minimal and absorptive for $\circ$.
3. $\mathcal{R} : B \to V$ (*ranking measure*) is a function satisfying $\mathcal{R}(0) = n, \mathcal{R}(1) = e$ and
   $\mathcal{R}(A \cup B) = \max_\ll\{\mathcal{R}(A), \mathcal{R}(B)\}$.
4. If $A = \cup\{P_i \mid i \in I\}$ and for all $i \in I$, $\mathcal{R}(P_i) = n$, then $\mathcal{R}(A) = n$ (*coherence*).

Observe that coherence is automatically satisfied for compact boolean algebras (i.e. where every covering of an event has a finite subcovering). What we still have to prove is that ranking measures really fit into our general quasi-measure framework.

**Theorem 3.1** Ranking-measure spaces are quasi-measure spaces and ranking algebras are hierarchical valuation algebras.

**Proof:** I.0, I.1 : obvious, I.2, I.3 : properties of $\max_{\ll}$, $\circ$ and $n$, I.4 : absorptiveness of $n$ together with the group ordering condition, I.5, I.6 : immediate. I.7 : by the group ordering condition, I.8 : take $w = v'$, I.9 : let's assume $n \underset{\ll}{} v \underset{\ll}{} v' \underset{\ll}{} e$. If $v = n$, we choose $w = n$. If $v \neq n$, we consider the corresponding full group (cf. 2.3). Here, we can find a $w$ with $v' \circ w = v$. If $e \ll w$, then $v' = v' \circ e \ll v' \circ w = v$, which contradicts our assumption. Hence $w \underset{\ll}{} e$. Ranking algebras are hierarchical because they verify $\underset{\ll}{} = \ll\ll$.

Ranking measures are the simplest instances of quasi-measures. For every ordered commutative group $G$, we can define a corresponding ranking algebra $\mathcal{V}(G)$. Notice that we can embed $\mathcal{V}(Z)$ in each nontrivial $\mathcal{V}(G)$ ($V \neq \{n, e\}$), $Z = (Int, +, <)$ being the ordered additive group of integers. Other natural structures are $\mathcal{V}(Q)$ and $\mathcal{V}(R)$, for $Q = (Rat, +, <)$, $R = (\mathbb{R}, +, <)$. Ranking measures do not necessarily satisfy $\mathcal{R}(\cup\{A_i \mid i \in I\}) = \sup_{\ll}\{\mathcal{R}(A_i) \mid i \in I\}$ for mutually disjoint $A_i$. In the classical probabilistic framework, this relation can only be violated for uncountable $I$. Here, it may fail for countable $I$. The boolean algebras are most of the time set algebras. In fact, we can often lift ranking measures to power set structures.

**Theorem 3.2** Let $(\mathcal{B}, \mathcal{R}, \mathcal{V})$ be a ranking measure space where $\mathcal{B}$ is a set algebra on $S$, the valuation ordering $\ll$ is complete (i.e. $\ll$-infima exist) and $n \ll \inf_{\ll}\{\mathcal{R}(P) \mid P \in \mathcal{B}, n \ll \mathcal{R}(P)\}$. Then there is a canonical extension $\mathcal{R}^*$ of $\mathcal{R}$ to the powerset structure $\mathcal{B}^* = (2^S, \cup, \cap, -, \emptyset, S)$ on $S$ s.t. $(\mathcal{B}^*, \mathcal{R}^*, \mathcal{V})$ is a ranking measure space and $\mathcal{R}^*$ is the maximal extension w.r.t. the $\ll$-induced pointwise ordering on ranking-measures from $\mathcal{B}^*$ to $\mathcal{V}$.

**Proof sketch:** $\mathcal{R}^*$ defined by $\mathcal{R}^*(A) = \inf_{\ll}\{\mathcal{R}(P) \mid A \subseteq P, P \in \mathcal{B}\}$ for $A \subseteq S$ is as desired.

In the literature, ranking-measure-like notions have been considered for different purposes. In [Weydert 91], we have provided a simple ranking measure semantics for default conditionals and comparative modalities (formalizing negligibility) together with a corresponding axiomatic characterization. In [Weydert 93], this approach has been extended to default quantifiers, i.e. genuine first-order contexts. A semantics for belief based on sets of ranking measures, different from what we discuss in this paper, can be found in [Weydert 94].

A very influential and sophisticated framework has been proposed by Spohn [88, 90]. His ordinal conditional functions (OCF) are intended to measure relative disbelief and model iterated belief changes. Their domains have the form $2^S \setminus \{\emptyset\}$ and the valuation structure is $(Ord, +', <')$, with $Ord$ being the class of ordinal numbers $\{0, 1, 2, \ldots \omega, \omega +' 1, \ldots, \omega +' \omega, \omega +' \omega +' 1, \ldots\}$, $+'$ the ordinal number addition and $<'$ the converse of the ordinal number ordering ($\ldots <' \omega <' \ldots <' 1 <' 0$). Because there is no absorptive value for $\emptyset$ and, in particular, because $+'$ is not commutative ($\omega +' 1 \neq \omega = 1 +' \omega$), his approach is not directly subsumed by ours. We could use another connective $+^\circ$ based on polynomial representations ($\alpha = a_n \omega^{\beta_n} +' \ldots +' a_1 \omega^{\beta_1} +' a_0$), which is commutative. But we still wouldn't get accessibility because there is no $x$ with $1 +^\circ x = \omega$. Of course, for Spohn's natural conditional functions (NCF), which take their values from $Nat$, this problem doesn't arise. Here, the valuation structure basically becomes $\mathcal{V}(Z)$. For many practical purposes, this might be enough. In [Goldszmidt, Pearl 92], for instance, this formalism has been used for modeling nonmonotonic inference with variable-strength defaults.

Possibility measures [Dubois, Prade 91] are another related and quite popular quasi-probabilistic tool for reasoning about uncertainty. Here, the domains are power sets and the valuation structure is of the form $([0, 1], <)$. Because conditional possibility measures are defined through min, stronger $\circ$-like connectives are not considered. Spohn's account and possibility theory both assume that their measures are fully additive, i.e. $\mathcal{R}(A) = \sup_{\ll}\{\mathcal{R}(\{a\}) \mid a \in A\}$. However, this shouldn't be a general postulate. If we want to avoid an infinitary form of the lottery paradox, we will need semi-quantitative measures $\mathcal{R}$ compatible with homogeneously small singletons, i.e. with $v \in V$ s.t. $\mathcal{R}(\{a\}) = v \ll e$ for all $a \in S$.

## 4 CUMULATIVE MEASURES

Probabilistic and ranking measures are situated at the extremes of the quasi-measure range. Both fairly popular, they are characterized by different strengths and weaknesses. Probabilities are well suited for a fine-grained, e.g. decision-theoretic analysis in finite or continuous contexts where precise numbers are available. For their semi-qualitative counterparts, quite the opposite is true. They are most appropriate when we are interested in a simple, cheap, rough, mainly number-free evaluation of plausibility, e.g. in default reasoning or (full) belief revision. When modeling plain belief, for instance, we want to avoid any confidence loss when conjoining two beliefs. Within the classical probabilistic framework, this can only be achieved by attributing measure zero to $\neg\varphi$ if $\varphi$ is believed. However, when we have to revise our beliefs based on new evidence supporting $\neg\varphi$ we get into trouble because subsets of a null set cannot be differentiated probabilistically. For ranking measures, this problem doesn't arise. Here, believing $\varphi$ means associating a non-maximal, but not necessarily minimal ranking measure value to $\neg\varphi$. This guarantees that beliefs are closed under conjunction. At the same time, Spohn-type conditionalization allows us to formulate reasonable revision procedures. Furthermore, we do no longer have to care about measurability constraints, the prohibition of uniform countable sample spaces and the representation of impossibility. On the other hand, everybody will agree that the representational and inferential power of ranking measures is rather restricted. In fact, it seems as if both formalisms were largely complementary. Consequently, we should try to combine them in a powerful framework sharing the basic advantages of both.

The major issue is to extend the usual probabilistic framework far enough to allow differentiation among and especially conditioning on sets of vanishing probability. The idea is to introduce a hierarchy of mainly classical valuation contexts such that from the perspective of any given level, lower ranks are considered negligible or irrelevant, i.e. would get probability zero within a standard interpretation. Our strategy is to use a ranking algebra $\mathcal{V} = (V, \circ, \ll)$ for describing the rough, global hierarchical structure and the real probabilistic valuation algebra $\mathcal{R}_+ = (\mathbb{R}_+, +, \times, 0, 1, <)$, i.e. the standard positive real number algebra, for the more conventional, fine-grained local structure. The main task is now to merge these structures in a suitable way. This is done by the following definition.

**Definition 4.1** Let $\mathcal{V}^\circ = (V^\circ, \max_\ll, \circ^\circ, n^\circ, e^\circ, \ll^\circ)$ be a ranking and $\mathcal{R}_+ = (\mathbb{R}_+, +, \times, 0, 1, <)$ be the real valuation algebra. Then we call $\mathcal{H}(\mathcal{V}^\circ, \mathcal{R}_+) = (H(\mathcal{V}^\circ, \mathcal{R}_+), \#, \circ, n, e, \ll)$ a *cumulative algebra* with *global structure* $\mathcal{V}^\circ$ and *local structure* $\mathcal{R}_+$ iff

- $H(\mathcal{V}^\circ, \mathcal{R}_+) = \{(n, 0)\} \cup V^\circ\setminus\{n\} \times \mathbb{R}_+\setminus\{0\}$, and for all $(a, r), (a', r') \in H(\mathcal{V}^\circ, \mathcal{R}_+)$,

- $(a, r) \# (a', r') = (a, r + r')$ if $a = a'$,
  $= (a', r')$ if $a \ll^\circ a'$,
  $= (a, r)$ if $a' \ll^\circ a$,

- $(a, r) \circ (a', r') = (a \circ^\circ a', r \times r')$,

- $n = (n^\circ, 0)$ and $e = (e^\circ, 1)$,

- $(a, r) \ll (a', r')$ iff $a \ll^\circ a'$ or $(a = a'$ and $a < a')$.

The definition of $H(\mathcal{V}^\circ, \mathcal{R}_+)$ as a proper subset of $V^\circ \times \mathbb{R}_+$ excluding $(n^\circ, r)$ for $r \neq 0$ and $(a, 0)$ for $a \neq n^\circ$ ensures that there are no zero-divisors ($x, y \neq n$ with $x \circ y = n$, e.g. for $x = (a, 0)$ and $y = (n^\circ, r)$) or trivial violations of multiplicative monotony by $(a, 0) \neq n$, $(e, r) \ll (e, 1)$ and $(a, 0) \circ (e, r) = (a, 0) \circ (e, 1)$. For the multiplicative connective, the componentwise strategy is the obvious one to guarantee those properties which are required for encoding independency (cf part 2). The lexicographic ordering reflects our basic stratification philosophy. Finding an appropriate additive connective, however, is less straightforward. Within single ranks ($a = a'$), we must adopt the pointwise approach $(a, r) \# (a', r') = (\max_\ll\{a, a'\}, r + r')$, because distributivity and the intended isomorphism between $\{e\} \times \mathbb{R}_+\setminus\{0\}$ and $\mathcal{R}_+\setminus\{0\}$ (i.e. real valuation algebra at the top) allow the derivation of $(a, r) \# (a, r') = ((a, 1) \circ (e, r)) \# ((a, 1) \circ (e, r')) = (a, 1) \circ ((e, r) \# (e, r')) = (a, 1) \circ (e, r + r') = (a, r + r')$. But additions involving different ranks cannot be handled that way. To satisfy additive accessibility, e.g. to pass from $(a, 1)$ with $a \ll^\circ e$ to the bigger value $(e, r)$ for $r < 1$, we would have to introduce negative numbers on the right-hand-side, which simply doesn't make sense. In fact, it would again bring in oddities like $(a, 0)$ for $a \neq n^\circ$. Notice that multiplicative accessibility already has brought us to consider $(a, r)$ with $1 < r$ to get from $(e, 1/r)$ to $(a, 1)$. Taking all these precautions, our structures are well-behaved.

**Theorem 4.1** Cumulative algebras are hierarchical valuation algebras. They verify **SH** if the ranking algebra is non-trivial ($V^\circ \neq \{n^\circ, e^\circ\}$).

**Proof :** Let $\mathcal{H}(\mathcal{V}^\circ, \mathcal{R}_+) = (H(\mathcal{V}^\circ, \mathcal{R}_+), \#, \circ, n, e, \ll)$ be a cumulative algebra.

1. $\mathcal{H}(\mathcal{V}^\circ, \mathcal{R}_+)$ is a valuation algebra.

**I.2** *Additive structure*. Commutativity and the neutrality of $n = (n^\circ, 0)$ are obvious. To see that associativity holds, consider any $((a, r) \# (a', r')) \# (a'', r'')$ and $(a, r) \# ((a', r') \# (a'', r''))$. If one left component is bigger than the other two, the sums are just identical to the corresponding pair. If the left components are equal, we can use the associativity of $+$. If only two of them are identical and the remaining one is smaller, we can drop the associated pair and the above expressions become equal to the sum of the former.

**I.3** *Multiplicative structure*. Obvious because of the corresponding properties for $(\mathcal{V}^\circ, \circ^\circ)$ and $(\mathbb{R}_+, \times)$.

**I.4** *Distributivity*. If $a' = a''$, then $(a, r) \circ ((a', r') \# (a'', r'')) = (a \circ^\circ a', r \times (r' + r'')) = (a \circ^\circ a', r \times r' + r \times r'') = (a \circ^\circ a', r \times r') \# (a \circ^\circ a'', r \times r'') = ((a, r) \circ (a', r')) \# ((a, r) \circ (a'', r''))$. If $a' \ll^\circ a''$, then $(a, r) \circ ((a', r') \# (a'', r'')) = (a, r) \circ (a'', r'') = ((a, r) \circ (a', r')) \# ((a, r) \circ (a'', r''))$ by the multiplicative monotony of $\circ^\circ$ or, if $a = n^\circ$ and $r = 0$, by absorption.

**I.5** *Linearity*. Follows directly from that of $\ll^\circ$ and $<$.

**I.6** *Additive monotony*. If $a \ll^\circ b \ll^\circ a'$, then $(a, r) \ll (a', r')$ implies $(a, r) \circ (b, s) = (b, s) \ll (a', r') = (a', r') \circ (b, s)$. The remaining possibilities are immediate.

**I.7** *Multiplicative monotony*. Immediate from the multiplicative monotony of $\mathcal{V}^\circ$ and $\mathcal{R}_+$ and $n = (n^\circ, 0)$.

**I.8** *Additive accessibility*. If $(a, r) \ll (a', r')$, then we have either $a \ll^\circ a'$ and $(a, r) \# (a', r') = (a', r')$, or $a = a'$ and $r < r'$ and the additive accessibility of $\mathcal{R}_+$ gives us $r''$ with $(a, r) \# (a, r'') = (a', r')$.

**I.9** Multiplicative accessibility. By the corresponding feature of $\mathcal{V}^\circ$ and $\mathcal{R}_+$ and the definition of $H(\mathcal{V}^\circ, \mathcal{R}_+)$.

2. $\mathcal{H}(\mathcal{V}^\circ, \mathcal{R}_+)$ is hierarchical. $(a, r) \ll\ll (a', r')$ iff $(a', r') \# (a, r) = (a', r')$ iff $a \ll^\circ a'$ or $(a, r) = (n^\circ, 0)$, which describes a modular relation.

3. **SH** holds. $(e^\circ, 1) \# (a, r) = (e^\circ, 1)$ for some $(a, r)$ with $(n^\circ, 0) \ll (a, r) \ll (e^\circ, 1)$, which exists by $\mathcal{V}^\circ$'s nontriviality.

The next definition now introduces our main concept.

**Definition 4.2** A quasi-measure whose valuation algebra is a cumulative algebra is called a *cumulative measure*.

Notice that real measures can be seen as special cumulative measures using the trivial ranking algebra, i.e. where $\mathcal{V}° = (\{n°, e°\}, ·°, \ll°)$. The main purpose of our cumulative constructions is to provide a general, quasi-probabilistic framework supporting something like shifting granularity. Locally, we are dealing with a classical numerical context, but globally, we adopt the ranking perspective. Conditioning allows us to pass between levels and, as we shall see, to model plain belief revision without being forced to give up all the useful classical probabilistic or measure-theoretic tools. The idea is to use the (top-ranked) local probabilities for decision-theoretic purposes and the global ranking structure for revision tasks. Of course, we have to expect that some features get lost. Completeness, for instance, can at best be guaranteed within bounded segments of ranks. That is, there might well be disjoint $A_i$, e.g. with $\mathcal{R}(A_i) = (a, i)$ for $i \in$ **Nat**, where $\sup_{\ll}\{\mathcal{R}(A_i) | i \in$ **Nat**$\}$ doesn't exist. On the other hand, we are no longer troubled by measurability constraints. Every cumulative (and therefore every probabilistic) measure on a set-algebra can be extended to a cumulative measure on the whole powerset.

**Theorem 4.2** Let $(\mathcal{B}, \mathcal{R}, H(\mathcal{V}, \mathcal{R}_+))$ be a cumulative measure space where $\mathcal{B}$ is a set algebra on S and $\mathcal{R}_+ = (\mathbb{R}_+, +, \times, 0, 1, <)$. Then there is an extension $\mathcal{R}^*$ of $\mathcal{R}$ to the powerset structure $\mathcal{B}^* = (2^S, \cup, \cap, -, \emptyset, S)$ on S and a ranking algebra $\mathcal{V}^*$ s.t. $\mathcal{R}^*$ is a cumulative measure and $\mathcal{V}$ a substructure of $\mathcal{V}^*$.

This result is different from our corresponding theorem 3.2 for ranking measures insofar as here, in general, we don't have neither a canonical extension nor constant valuation algebras (i.e. $\mathcal{V} = \mathcal{V}'$).

Another proposal to reconcile the advantages of probabilistic and ranking measures has been made by Boutilier [Boutilier 93]. Roughly speaking, the structures he considers could be interpreted as special, restricted instances of cumulative measure spaces. Basically, his approach has been designed for finite boolean algebras, but - to a certain extent - it can be generalized to infinite ones. The idea is to have a possibility ranking on atoms and a probability measure $\mathcal{P}_r$ attached to each level r which vanishes for measurable propositions built up from atoms of other ranks. Given an event A, we look for the highest rank r where its probability $\mathcal{P}_r(A)$ becomes strictly positive. Boutilier doesn't present a fully integrated account with combined valuation scales, but in our terminology, A would get the value $(r, \mathcal{P}_r(A))$.

The approach, as it stands, has several shortcomings. First of all, it relies on possibility measures, which are not general enough for some purposes (e.g. for modeling infinite boolean algebras based on countably many small atoms). Secondly, his definition forces us to use well-founded rankings, which seems to be a quite artificial restriction. In fact, the only well-founded ranking algebras are $\mathcal{V}(Z)$ and the trivial one. But the main problem with Boutilier's theory is that it doesn't really fit into the quasi-measure framework. Because the local values are supplied by probabilistic valuations, they cannot exceed 1. Consequently, multiplicative accessibility will fail, i.e. there will be no smooth account of conditional measures and our valuation freedom will be severely restricted. For instance, we may encounter valuated events, less "probable" than their complements, for which there can't be another event which is exactly two times as "probable". So, it seems justified to say that Boutilier's formalism achieves only a partial integration of the quantitative and the qualitative viewpoint.

## 5 BELIEF STRUCTURES

Traditionally, epistemic states of ideal agents have often been represented or approximated by probabilistic or discrete ranking measures on world sets. Unfortunately, as mentioned before, these pure approaches exhibit serious weaknesses. The first one, for instance, fails either to represent (by admitting full beliefs with probabilities below 1) or to revise plain belief (trivialization by conditioning on null sets). It also runs into problems because of measurability constraints affecting the realization of countably infinite uniform belief spaces. The second one, on the other hand, is unsuitable for fine-grained representation and decision-taking tasks where it makes a difference whether one option is as or twice as plausible as another one (which cannot be expressed by ranking measures). In particular, there is no possibility to model statistical knowledge directly by Spohn-type degrees of belief. Basically, these problems call for refined valuation structures backing the necessary differentiations. Of course, this can be done in many different ways. Either indirectly and hidden by a sophisticated machinery [Bacchus et al. 92], or directly, by combining complementary uncertainty formalisms, as we do. In the following, we are going to sketch some ingredients of an epistemic framework based on cumulative measures, which allows a correct handling of our benchmarks.

First of all, we have to address the question about the nature of epistemic states. Here, we want to adopt a liberal attitude. Epistemic states should not be identified with any kind of measures. This would be an oversimplification contradicting our subjective experience of what constitutes real-world belief (inconsistency, partiality, vagueness, mixture of qualitative and quantitative contents, deductive incompleteness, ... ). What we will assume, however, is that any given epistemic state s can be interpreted or evaluated so as to provide a cumulative measure $\mathcal{R}s$ describing the corresponding official, surface belief structure, to be exploited for practical purposes like decision-tasks. This approach might be called the *projection model of belief*. The domain of $\mathcal{R}s$ is formed by the propositions of our actual language (state) $\mathcal{L}s$, which we assume to be boolean. The valuation algebra is assumed to be non-trivial. Backed by the pleasant properties of cumulative measures, we can now represent plain belief in the traditional way without having to care about the shortcomings of the probabilistic account. Let's express "*within the epistemic state* s, A *is plainly believed*" by $s \models B(A)$.

**Plain belief :**

   $s \models B(A)$ iff $\mathcal{R}s(A) = (e°, 1)$

What's important is that this doesn't force us to stipulate $\mathcal{R}s(\neg A) = (n°, 0)$, i.e. the impossibility of $\neg A$, because $(e°, 1) = \mathcal{R}s(T) = \mathcal{R}s(A \vee \neg A) = \mathcal{R}s(A) \# \mathcal{R}s(\neg A) = (e°, 1) \# (a, r)$ is possible and true for $(n°, 0) \ll (a, r) \ll (e°, 1)$. Plainly believing A is not the same as rejecting the possibility of $\neg A$. Also, we can differentiate between beliefs with regard to their revisability or epistemic entrenchment by attributing either bigger or smaller measure values to their complements (cf. below). A basic characteristic of plain beliefs is their closure under conjunctions.

**Conjunctive closure :**

   If $\mathcal{R}s(A), \mathcal{R}s(A') = (e°, 1)$, then $\mathcal{R}s(A \& A') = (e°, 1)$

This holds because $\mathcal{R}s(\neg A \vee \neg A') = \max_{\ll} \{\mathcal{R}s(\neg A), \mathcal{R}s(\neg A')\} \ll\ll (e°, 1)$. Of course, we may also adopt less certain attitudes and assign intermediate values. Indifference w.r.t. A, for instance, amounts to state $\mathcal{R}s(\neg A) = \mathcal{R}s(A) = (e°, 0.5)$. In first-order contexts, it is easy to imagine situations where the additive structure of classical valuation algebras becomes inappropriate. For instance, let $\Psi(x)$ be an opaque number-theoretic property for which we know that there is exactly one solution, but where we cannot give any bounds. Then, it seems reasonable to adopt a symmetric attitude, which can easily be realized within our cumulative framework. Let's assume that $Ls$ is powerful enough to express $\Psi(\underline{n})$, the $\underline{n}$ being natural number constants.

**Infinitary uniformity :**

   $\mathcal{R}s(\Psi(\underline{0})) = \mathcal{R}s(\Psi(\underline{1})) = \ldots = (a, 1) \ll\ll (e°, 1)$, e.g.

In addition, we need of course a possibly nondeterministic revision mechanism $\mathcal{N}$ which, given an epistemic state $s$ and a new item of information $i$, defines a set of possible updates $\mathcal{N}[i](s)$. In general, $i$ will be a constraint for the cumulative measures $\mathcal{R}s'$ of admissible revision states $s'$, e.g. $i = [\mathcal{R}(A) = (e°, 1)]$. A minimal condition for this $\mathcal{N}[i]$ can be borrowed from the classical paradigm.

**Top-conditionalization :**

   $\mathcal{R}s'(B) = \mathcal{R}s(B \mid A)$ if $\mathcal{R}s(B \mid A)$ has the form $(e°, r)$

We cannot require full conditionalization because this would force us to accept $\mathcal{R}s'(\neg A) = \mathcal{R}s(\neg A \mid A) = (n°, 0)$, precluding nontrivial updating with $\neg A$. That is, we would be confronted again to the problems of the traditional framework, we want to escape.

**Acknowledgements**

This work was supported by the German Ministery of Research and Technology under grant ITS 9102, project LOGO. Thanks to the anonymous referees, whose comments have helped to make this paper at least slightly more intelligible.